\documentclass[10pt,twocolumn,letterpaper]{article}

\usepackage[pagenumbers]{cvpr} 

\newcommand{\ourmodel}{TFDM}
\usepackage{bbm}
\usepackage{microtype}

\definecolor{cvprblue}{rgb}{0.21,0.49,0.74}
\usepackage[pagebackref,breaklinks,colorlinks,allcolors=cvprblue]
{hyperref}
\usepackage{amsmath}
\usepackage{amssymb}
\usepackage{mathabx}
\usepackage{tabularx,booktabs}
\usepackage{graphicx}
\usepackage{multirow}
\usepackage{colortbl}
\usepackage[normalem]{ulem}

\def\paperID{3955\vspace{-6mm}} 
\def\confName{ICCV\vspace{-4pt}}
\def\confYear{2025}

\title{\vspace{-1.3cm}{\ourmodel}: Time-Variant Frequency-Based Point Cloud Diffusion with Mamba \vspace{-0.5cm}}

\author{%
Jiaxu Liu \\
Durham University \\
{\tt\small jiaxu.liu@durham.ac.uk}
\and
Li Li \\
King's College London \\
{\tt\small li.8.li@kcl.ac.uk}
\and
Hubert P. H. Shum \\
Durham University \\
{\tt\small hubert.shum@durham.ac.uk}
\and
Toby P. Breckon \\
Durham University \\
{\tt\small toby.breckon@durham.ac.uk}
}

\begin{document}
\maketitle
\definecolor{notecolor}{rgb}{0.1,0.1,0.9}
\newcommand{\notes}[1]{{\textcolor{notecolor}{#1}}}
\newcommand{\minor}[1]{{\textcolor{brown}{#1}}}
\newcommand{\luis}[1]{{\textcolor{cvprblue}{[Luis] #1}}}
\newcommand\sgm[1] {\notes{#1}}
\begin{abstract}
Diffusion models currently demonstrate impressive performance over various generative tasks. Recent work on image diffusion highlights the strong capabilities of Mamba (state space models) due to its efficient handling of long-range dependencies and sequential data modeling. Unfortunately, joint consideration of state space models with 3D point cloud generation remains limited. To harness the powerful capabilities of the Mamba model for 3D point cloud generation, we propose a novel diffusion framework containing dual latent Mamba block (DM-Block) and a time-variant frequency encoder (TF-Encoder). The DM-Block apply a space-filling curve to reorder points into sequences suitable for Mamba state-space modeling, while operating in a latent space to mitigate the computational overhead that arises from direct 3D data processing. Meanwhile, the TF-Encoder takes advantage of the ability of the diffusion model to refine fine details in later recovery stages by prioritizing key points within the U-Net architecture. This frequency-based mechanism ensures enhanced detail quality in the final stages of generation. Experimental results on the ShapeNet-v2 dataset demonstrate that our method achieves state-of-the-art performance (ShapeNet-v2: 0.14\% on 1-NNA-Abs50 EMD and 57.90\% on COV EMD) on certain metrics for specific categories while reducing computational parameters and inference time by up to 10$\times$ and 9$\times$, respectively. Source code is available in Supplementary Materials and will be released soon.
\end{abstract}





\section{Introduction}
\label{sec:intro}
Point clouds have emerged as a powerful representation in 3D data processing due to their high fidelity, ease of capture, and straightforward manipulation. 3D point cloud generation has gained increasing attention for its superior performance across various applications, including virtual reality, robotics~\cite{robotics}, mesh generation, scene completion, and reconstruction~\cite{scene_completion,mesh_gen,reconstr_diffusion}. Despite advancements in 2D image generation~\cite{ddpm_first_2d, ddpm_initial-idea}, point clouds remain inherently discrete, unordered, and complex, presenting unique challenges that require further exploration to facilitate effective use with contemporary generative models.

Existing generative models targeting point clouds span a wide range of methods, including variational autoencoders (VAE)~\cite{pc_vae_setvae}, generative adversarial networks (GAN)~\cite{pc_GANs_rGAN,pc_GANs_shapeGF}, and normalizing flows~\cite{pc_flow_DPF-Net,pc_flow_pointflow}. 
However, these methods often face challenges in achieving stable and high-fidelity generation, limiting their effectiveness in complex 3D point cloud tasks. Recently, denoising diffusion models~\cite{pc_ddpm_dpm} have demonstrated superior 3D point cloud generation performance, by defining a forward process that gradually perturbs the point cloud into standard Gaussian noise, and then learns to recover that original cloud through a reverse denoising process. Once trained, new point clouds can be generated by directly sampling from the Gaussian distribution and using the same progressive denoising process, offering a more robust and accurate approach to 3D point cloud generation. Despite these advantages, the complexity of diffusion models imposes high computational demands, making scalability challenging for efficient point cloud generation.

\begin{figure}[t]
\begin{center}
\includegraphics[width=1.0\linewidth]{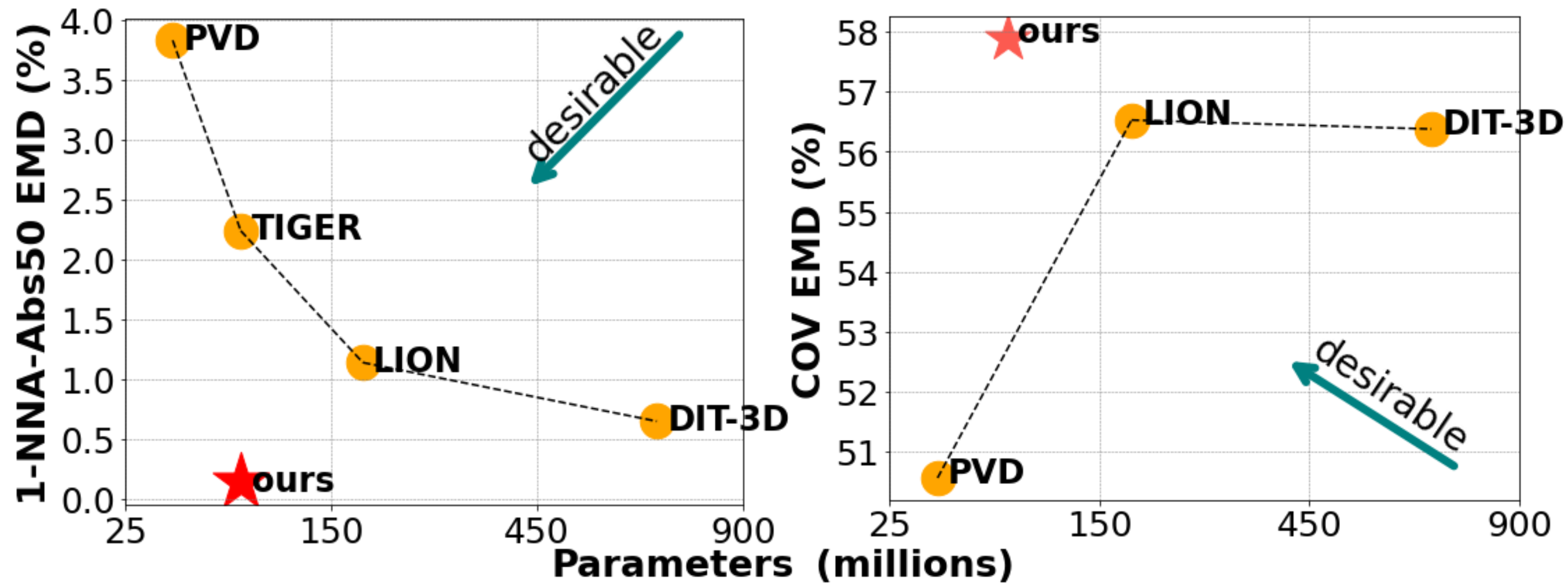}

\end{center}
   \vspace{-12pt}
   \caption{1-NNA-Abs50 EMD \& COV EMD (~\cref{sec:exp-1}) performance (\%)  \textit{ vs.} parameter size (millions) on ShapeNet-v2 Car category. For 1-NNA-Abs50 EMD (left), lower value indicates better generation quality and fidelity. For COV EMD (right), higher is better diversity. In both plots, moving left along the horizontal axis denotes smaller model.}
   

   \vspace{-4mm}
\label{fig:start}
\end{figure}

Recent advancements in sequence modeling techniques, particularly the Mamba architecture~\cite{mamba_initial} leveraging state space models for effective sequence handling, have demonstrated significant potential. Previous studies~\cite{pointmamba_point_cloud,pc-point-cloud-mamba} have shown the superiority of Mamba in downstream point cloud tasks, highlighting its effectiveness. Meanwhile, diffusion models have also shown promise in point cloud processing~\cite{pc_ddpm_dpm,pc_ddpm_pvd}. Recent research has explored transformer-based architectures~\cite{pc_ddpm_tiger,pc_ddpm_DIT-3D} to enhance the network design, while others have leveraged latent spaces~\cite{pc_ddpm_freq,pc_ddpm_lion} to improve performance. However, diffusion models require substantial computational resources, particularly due to the multiple iterations involved in the reverse process, which makes both training and inference highly time-consuming. These demands are especially prohibitive in real-time or resource-constrained environments. Whilst Mamba is more computationally lightweight than a transformer architecture, its integration with diffusion processes for 3D point clouds remains underexplored due to these computational challenges.

Several recent works have focused on frequency analysis~\cite{gda,awt,ddpm_2d_freq1,ddpm_2d_freq2} in both 2D and 3D contexts. Among these, some studies have explored the potential of integrating frequency analysis with diffusion models~\cite{2D_ddpm_slim,ddpm_2d_freq1} and Mamba architectures on 2D images~\cite{ddpm_2d_zigma}. Additionally, very recent work~\cite{freq-mam-dif} combines frequency analysis and Mamba within diffusion models in 2D. However, applying frequency analysis to 3D point clouds presents unique challenges. In 2D, continuous frequency decomposition can be readily achieved via Fourier transforms or other spectral analysis methods. In contrast, point clouds are inherently discrete and sparse, making it significantly more challenging to integrate frequency analysis with state-space models in 3D.

To address these aforementioned research gaps, we propose {\ourmodel}, a novel point cloud diffusion architecture integrated with the Mamba framework. Given the high computational and memory demands of modeling long sequences, we improve efficiency by introducing dual latent Mamba blocks (DM-Block) in the latent space. This novel design reduces model size while preserving performance, offering a more compact yet effective approach compared to conventional Mamba-based architectures. Furthermore, we draw inspiration from 2D image diffusion, where coarse structures (low-frequency components) are recovered at an early stage with refined details (high-frequency components) recovered later. This pattern also extends to 3D point clouds: the generation process initially forms a blurred overall shape before refining contours, where high-frequency components are related to edges and corners, and flat regions to low-frequency features. Based on this observation, we emphasize high-frequency areas in later time steps. Specifically, by employing a U-Net architecture with multiple downsampling layers, we propose time-variant frequency-based encoder (TF-Encoder), replacing traditional farthest point sampling with our frequency-based method to better select key points in later time steps, thereby capturing more details during the final recovery stages. 

Overall, our contributions can be summarized as follows:

\begin{itemize} 
    \item The first joint use of frequency-based analysis for Denoising Diffusion Probabilistic Models (DDPM) combined with the use of Mamba architecture, to address the computational demands of 3D point cloud diffusion modeling.
    \item A novel end-to-end architecture (\ourmodel) that integrates a promising \textbf{t}ime-variant \textbf{f}requency-based \textbf{e}ncoder (TF-Encoder) with \textbf{d}ual latent \textbf{m}amba \textbf{b}lock (DM-Block) to enhance high-frequency point cloud details. It adapts to the diffusion timestep within the Mamba latent space of a point cloud DDPM, ensuring precise detail refinement.
    \item Extensive experiments on the established ShapeNet-v2~\cite{chang2015shapenet} benchmark dataset that demonstrates both state-of-the-art (SoTA) performance (ShapeNet-v2: 0.14\% on 1-NNA-Abs50 EMD and 57.90\% on COV EMD) and the efficacy (reducing up to 10$\times$ and 9$\times$ on parameters and inference time) of our approach across multiple reference categories. 
\end{itemize}

\vspace{-0.2cm}
\section{Related Work}
\label{sec:related work}

We briefly present relevant prior work spanning Diffusion, Point Cloud Generation, State Space Models, and Frequency Analysis).

\noindent
\textbf{Diffusion Models:} Denoising diffusion probabilistic models (DDPM)~\cite{ddpm_initial-idea,ddpm_first_2d} generate data by reversing a progressive noising process in a Markov chain, offering stable training and robust sample fidelity. They avoid typical pitfalls such as mode collapse seen in other generative frameworks~\cite{pc_GANs_shapeGF,pc_vae_setvae}. Beyond 2D applications~\cite{ddpm_initial-idea,ddpm_first_2d}, several studies have adapted diffusion models for 3D point clouds~\cite{pc_ddpm_lion,pc_ddpm_DIT-3D,pc_ddpm_tiger}. Despite promising results, these approaches often incur high computational overhead due to iterative denoising steps, especially when dealing with large 3D datasets. Our work addresses this efficiency gap via a latent-space design that reduces computational cost without sacrificing generative quality.


\begin{figure*}[hbt!]
\begin{center}
\includegraphics[width=1.04\linewidth]{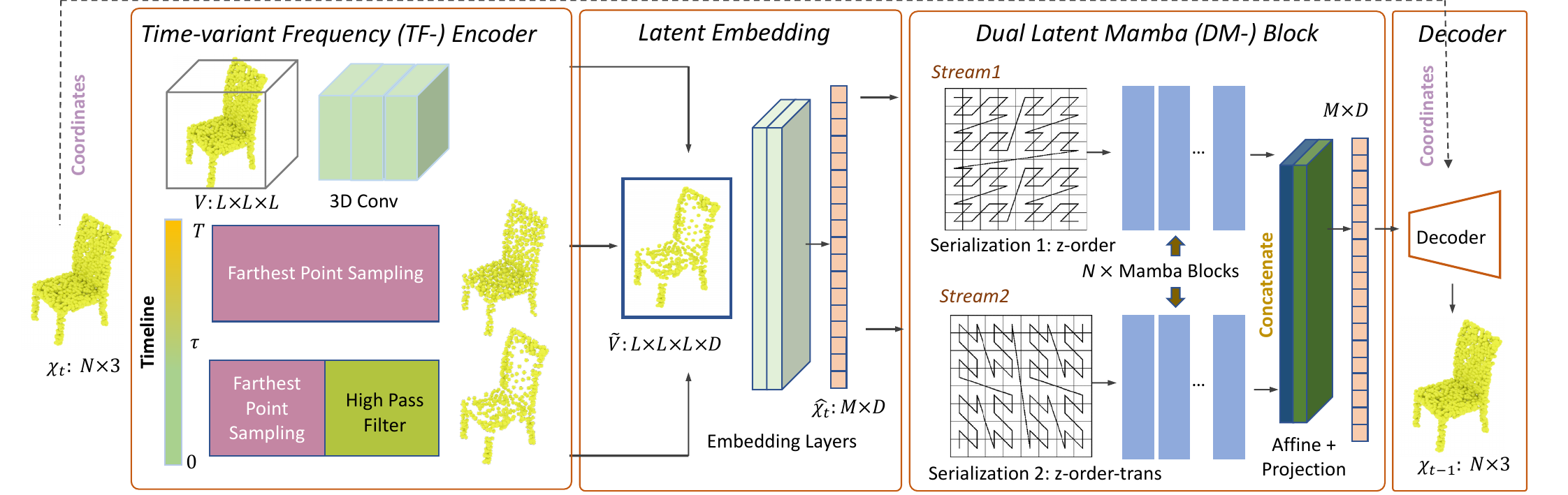} 
\end{center}
   \vspace{-3mm}
   \caption{The overview architecture of our proposed {\ourmodel}. The network takes a point cloud at timestep $t$ as input and aims to predict the noise component in $\mathcal{X}_t$ to obtain the point cloud at timestep $t-1$. Initially, the input point cloud is passed through a time-variant frequency-based encoder. This is followed by a latent embedding module that generates a latent point cloud $\hat{\mathcal{X}_t}$. The latent point cloud is then processed through Two-Streams Mamba blocks, which apply different serialization methods to extract diverse and complementary features. Subsequently, an affine transformation block is employed to align the latent point clouds from the different streams, ensuring consistency and integration of the extracted features. Finally, the aligned latent representation is decoded back into the 3D space.}
   \vspace{-3mm}
\label{fig:overview}
\end{figure*}

\noindent
\textls[-8]{\textbf{Point Cloud Generation:} Point cloud generation has been explored through various paradigms, including auto-regressive models~\cite{pc_AR_2, pc_AR_can}, flow-based techniques~\cite{pc_flow_DPF-Net, pc_flow_softflow, pc_flow_pointflow}, and GAN~\cite{pc_GANs_rGAN, pc_GANs_shapeGF, chang2015shapenet}. More recent efforts employ diffusion for 3D shapes~\cite{pc_ddpm_DIT-3D,pc_ddpm_tiger,pc_ddpm_lion} incorporating techniques such as merging point and voxel representations~\cite{pc_ddpm_pvd}, constructing hierarchical latent spaces~\cite{pc_ddpm_lion} to model high-level shape semantics, integrating frequency-based loss frequency-based loss function~\cite{pc_ddpm_freq}, and adapted Transformers~\cite{pc_ddpm_tiger,pc_ddpm_DIT-3D}. Although these works have advanced point cloud fidelity, many still struggle with high computational costs in capturing 3D structures. We introduce a more efficient method that operates in the latent space yet retains the ability to model fine geometric details.}

\noindent
\textls[-5]{\textbf{State Space Model:} State space models~\cite{mamba_initial,mamba_s4} leverage linear recurrences to handle long-range dependencies with fewer parameters. ~\cite{mamba_s4} notably improved sequence modeling in various tasks, and Mamba~\cite{mamba_initial} further enhanced efficiency by mitigating linear-time inference overhead. While Mamba has been applied to point cloud tasks such as classification and segmentation~\cite{pointmamba_point_cloud,pc-point-cloud-mamba} or even 2D image diffusion~\cite{ddpm_2d_zigma,ddpm_2d_mamba}, its use in 3D diffusion remains largely underexplored. We bridge this gap by employing Mamba as the core sequence modeling engine in our diffusion framework, tailored specifically for point cloud generation.}


\noindent
\textbf{Frequency Analysis:} Frequency decomposition methods have proven effective in identifying high-frequency (edge-like) vs. low-frequency (smooth) geometry, both for point clouds~\cite{gda,awt} and images~\cite{2D_ddpm_slim}. Recent diffusion-based studies~\cite{2D_Wave_Fast_Saclable_Diff,ddpm_2d_freq1} exploit wavelet or spectral transforms to boost detail recovery and reduce redundancy. Li et al.~\cite{pc_ddpm_freq} introduced a frequency-oriented module to refine 3D point cloud diffusion but did not pair it with state space modeling. In contrast, we integrate frequency-aware sampling into our Mamba-augmented diffusion approach, significantly improving detail fidelity while remaining computationally efficient.


\section{Methodology}
\label{sec:method}
We begin by formulating the generative diffusion objective in ~\cref{sec:4.1}. Building on the robust 3D modeling capacity of Mamba blocks, we propose a novel diffusion framework for point clouds (see ~\cref{fig:overview}). Specifically, in ~\cref{sec:4.2} we introduce a frequency-based point cloud filter to extract key frequency components. In ~\cref{sec:4.3}, we describe a time-variant frequency encoder that uses these components for key-point sampling. Finally, in ~\cref{sec:4.4}, we present our two-stream latent Mamba architecture, which integrates state space modeling, frequency analysis, and diffusion to generate high-fidelity point clouds efficiently.

\begin{figure*}[hbt!]
\begin{center}
\includegraphics[width=1.0\linewidth]{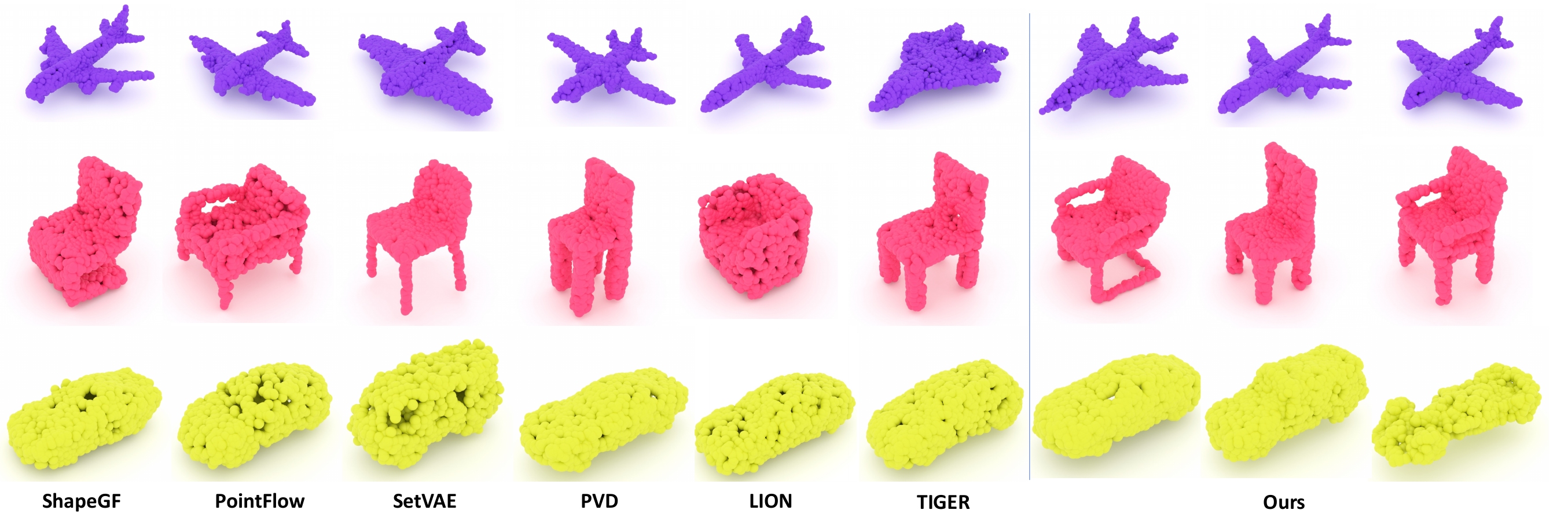} 
\end{center}
   \vspace{-3mm}
   \caption{Qualitative results comparing our approach (right) with other leading contemporary approaches (left/middle), our {\ourmodel} can generate high-quality and diverse point clouds. Three illustrative object categories $\{airplanes, chairs, cars\}$ are included here only.}
   \vspace{-3mm}
\label{fig:qualitative_results}
\end{figure*}
\subsection{Generative Modeling of Point Clouds}
\label{sec:4.1}
Given a point cloud $\mathcal{X} \in \mathbb{R}^{N \times 3}$ consisting of $N$ points, our goal is to generate a high-fidelity point cloud from Gaussian noise $p(\mathbf{x}_{T})$ by learning the transition probability $p_{\theta}(\mathbf{x}_{t-1}|\mathbf{x}_{t})$. Specifically, we model the mean of the transition distribution while keeping a predetermined variance throughout the diffusion reverse process. Similar to TIGER~\cite{pc_ddpm_tiger} and PVD~\cite{pc_ddpm_pvd}, we employ a U-Net backbone for ${\mu}_\theta(\mathbf{x}_t, t)$ to incorporating a newly designed Mamba layer and frequency-based key point selection to enhance its capability. To sample the point cloud, we perform denoising from $p(\mathbf{x}_{T})$ over $T$ timesteps by minimizing the MSE discrepancy loss $\mathbb{E}_{t\sim [1,T]}\parallel{{\mathbf{\epsilon}}}_\theta\left(\mathbf{x}_{t},t\right)-{\epsilon}_0\parallel_2^2$ between the predicted noise $\epsilon_\theta(\mathbf{x}_t, t)$ and the true noise $\epsilon_0$, ensuring accurate denoising performance across different time steps.

\subsection{Point Cloud Graph Filter}
\label{sec:4.2}
Unlike the 2D domain, where spectral analysis methods such as Fourier and wavelet transforms are straightforwardly applicable~\cite{awt,gda,2D_Wave_Fast_Saclable_Diff}, the irregular, non-Euclidean nature of point clouds \cite{pc-unorder-structure,pc-unorder-structure2} demands the development of alternative approaches for defining frequency components. The absence of a structured grid in point cloud data \cite{pc-unorder-structure3,point_transformer_v1} complicates the direct adoption of traditional spectral techniques, thus motivating a tailored method to effectively capture and process the inherent frequency characteristics of point cloud geometry.

\vspace{4pt}
\noindent
\textbf{Graph Construction}: 
To capture the geometric structure in point clouds and topological relationships between points, we construct a $k$-nearest neighbors (k-NN) graph. The graph signals are further leveraged to extract high-frequency points with no trainable parameters.

Given a point cloud $\mathcal{X} = \{{x}_i , \ldots, {x}_N\}$ with corresponding $d$-dimensional features $\mathbf{f}_i \in \mathbb{R}^d$, $i \in \{1, \ldots, N\}$, we construct a $k$-NN graph $\mathcal{G} = (\mathcal{V}, \tilde{\mathcal{A}}_u, \tilde{\mathcal{A}}_w)$. Each point ${x}_i$ corresponds to a node $v_i \in \mathcal{V}$, $\tilde{\mathcal{A}}_u$ and $\tilde{\mathcal{A}}_w \in \mathbb{R}^{N \times N}$ are normalized unweighted and weighted adjacency matrices encoding point dependency in feature space. The unweighted $\tilde{\mathcal{A}}_{ij}^{u}$ and weighted edges $\tilde{\mathcal{A}}_{ij}^{w}$ connecting two nodes $v_i$ and $v_j$ are defined as:
\begin{align}
    \tilde{\mathcal{A}}_{ij}^{u} &= \mathbbm{1}({x}_j \in \mathcal{N}({x}_i)), \notag \\
    \tilde{\mathcal{A}}_{ij}^{w} &= \kappa(\|{x}_i - {x}_j\|^2) \cdot \tilde{\mathcal{A}}_{ij}^{u},
\end{align}
where $\kappa(\cdot)$ is a non-negative function, e.g., a Gaussian function, to ensure that $\tilde{\mathcal{A}}_w$ is a diagonally dominant matrix; $\mathcal{N}$ represents the neighborhood; $\mathbbm{1}(\cdot)$ represents the indicator function which returns 1 if the specified condition (the function input) is true and 0 if it is false.

\vspace{4pt}
\noindent
\textbf{Point Cloud High-pass Filter}:
\textls[-2]{Our design of the point cloud high-pass graph filter draws insights from the 2D case, where high-frequency components, corresponding to sharp pixel variations like edges, elicit strong responses in the spatial domain.
Following GDA~\cite{gda}, we construct our graph filter with the commonly adopted filter operator, specifically the Laplacian operator:
$h(\tilde{\mathcal{A}}_w) = I - \tilde{\mathcal{A}}_w$. It takes a graph signal $\mathbf{s}_d \in \mathbb{R}^N, \forall d \in \{1, \ldots, D\}$ and produce filtered $\mathbf{y}_d = h(\tilde{\mathcal{A}}_w) \cdot \mathbf{s}_d \in \mathbb{R}^N$, then the frequency response of $h(\tilde{\mathcal{A}}_w)$ with associated $\lambda_i$ is: }
\begin{equation}
    \widehat{h}(\tilde{\mathcal{A}}_w) = \text{diag}(1 - \tilde{\lambda}_1, 1 - \tilde{\lambda}_2, \dots, 1 - \tilde{\lambda}_N),
\end{equation}
where $\text{diag}(\cdot)$ denotes the diagonal matrix operator. The eigenvalues $\tilde{\lambda}_i$ are thus ordered reversely which represent the frequencies descending. As a result of the frequency response $1 - \tilde{\lambda}_i < 1 - \tilde{\lambda}_{i+1}$, the low frequency will be weakened which makes this to be a high-pass filter. 

We apply the filter $h(\tilde{\mathcal{A}}_w)$ to the point cloud $\mathcal{X}$ to obtain the filtered point $h(\tilde{\mathcal{A}}_w)\mathcal{X}$, with each point computed as:
\begin{equation}
(h(\tilde{\mathcal{A}}_w)\mathcal{X})_i = x_i - \sum_{j}^{N} (\tilde{\mathcal{A}}_w)_{i,j}x_j.
\label{eq:high_pass_filter}
\end{equation}
\textls[-13]{It preserves the variation information with neighbors, as the filtered point in~\cref{eq:high_pass_filter} computes the difference between a point feature and the linear combination of its neighbor features.}

Finally, we derive a frequency-based ordering for each point by computing the $l_2$-norm in Eq. (11). The top $M$ points are then selected to capture the most dominant high-frequency components. This approach effectively integrates frequency decomposition into the point cloud domain, despite its inherent irregularity.

\begin{figure}[t]
\begin{center}
\includegraphics[width=1.0\linewidth]{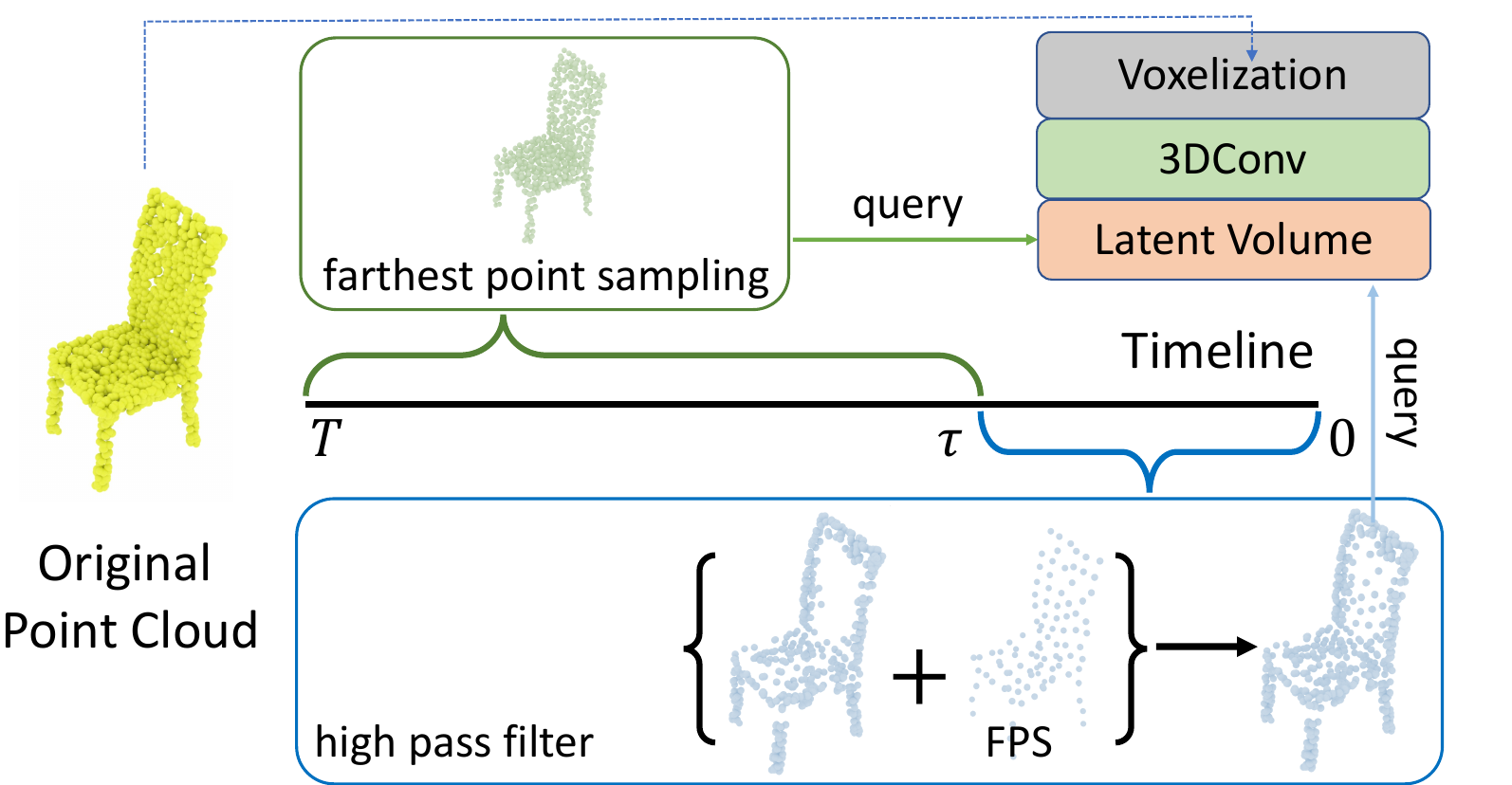}
\end{center}
   \vspace{-4pt}
   \caption{\textls[0]{Illustration of the frequency key point selection process within the encoder to show how different strategies are applied across various timelines to obtain a downsampled point cloud. Subsequently, the downsampled point cloud is used to query the latent volume, resulting in the latent point cloud.}}
\label{fig:encoder}
\end{figure}

\subsection{Time-Variant Frequency Point Cloud Encoder}
\label{sec:4.3}


As shown in~\cref{fig:encoder}, we introduce TF-Encoder, a novel encoding mechanism designed to fully leverage the timestep-dependent recovery dynamics of the diffusion process. Unlike static sampling strategies, TF-Encoder adaptively refines point cloud representations by selectively emphasizing frequency information at different timesteps. The core insight is that diffusion first reconstructs coarse, low-frequency structures and progressively refines high-frequency details in later stages. To align with this progression, TF-Encoder dynamically adjusts the sampling process, allocating a greater “high-frequency budget” to later timesteps, where local details become most critical.

\noindent
\textbf{Voxel-Based Feature Extraction}:
Specifically, we utilize the PVCNN~\cite{pvcnn} backbone, which enables efficient computation by downsampling the point cloud into a voxel grid. For a point cloud at any time step $t$, denoted as $\mathcal{X}_t \in \mathbb{R}^{N \times 3}$, our TF-E $\mathcal{E}$ transforms the point cloud into a latent space $\mathcal{\hat{X}}_t \in \mathbb{R}^{M \times D}$, where $M < N$, representing the number of subsampled and original points, respectively. To aggregate the voxelized features, the point cloud $\mathcal{X}_t$ with normalized coordinates ${c} = \left( {x},{y},{z} \right)\ $ need to be voxelized into the voxel grids $\left\{\boldsymbol{V}_{m,p,q}\right\}$, where $\boldsymbol{V} \in \mathbb{R}^{L \times L\times L}$ with resolution $L$. The interpolated latent feature $f_i$ for each voxel grid is computed as the mean of the features of the points within that grid: 
\begin{equation}
\begin{split}
    \boldsymbol{V}_{m,p,q} = \frac{1}{K_{m,p,q}} \sum_{i=1}^{n}\boldsymbol{I} [ \text{floor}({x_{i}}\times {r}) = m,\\ \text{floor}({y_{i}}\times {r}) = p,\ \text{floor}({z_{i}}\times {r}) = q ]\times {{f}_{i}},
\end{split}
\end{equation}
\textls[-10]{where $r$ denotes the voxel resolution and $\boldsymbol{I}$ is an indicator function that indicates whether coordinates $c_{i}$ belong to the voxel grid ${ \left\{ m,p,q \right\} }$. ${K_{m,p,q}}$ represents the count of points falling within the grid ${\left\{ m,p,q \right\}}$, and $\text{floor}(\cdot)$ is the floor function that outputs the greatest integer less than or equal to the input. After voxelization, multiple 3D convolutional layers with Swish activation~\cite{swish} and GroupNorm~\cite{group_norm} are applied to obtain the latent volume $\tilde{\boldsymbol{V}} \in \mathbb{R}^{L \times L\times L \times D}$ with $D$ channels.}

\noindent 
\textbf{Time-Variant Frequency-Aware Sampling }:
Unlike standard furthest point sampling (FPS) pipelines in the PVCNN backbone, we jointly incorporate a high-pass graph filter (\ref{sec:4.2}) with FPS in a time-variant manner. This design ensures that in the early time-steps $t<\tau$, we maintained a balanced selection of low-frequency structures and a subset of high-frequency regions for better global shape alignment. As the process advances to later time-steps $t\geq\tau$, our approach prioritizes high-frequency points, enabling the precise capture of subtle edges, corners, and intricate contours.


Formally, for $M$ target points, we select $\zeta M$ points with our graph-based high-pass filter, while the remaining $(1 - \zeta) M$ points are sampled via FPS. As  diffusion progresses, $\zeta$ can be adjusted to emphasize high-frequency details. This adaptive strategy allows TF-Encoder to align with the diffusion trajectory, ensuring time-specific frequency emphasis. The extracted point cloud is given by:
\begin{equation}
{
\mathcal{X}_t^{*}= 
\begin{cases} 
\zeta h(\tilde{\mathcal{A}}_w)\mathcal{X}_t + (1-\zeta)F(\mathcal{X}_t^{N-M})), & t = 0,1,...,\tau \\
F(\mathcal{X}_t), & t = \tau,...,T
\end{cases}}
\label{eq:subsample}
\end{equation}
where $F(\cdot)$ represents the furthest point sampling and $\mathcal{X}_t^{N-M}$ denotes the original point cloud excluding the points that passed the high-pass filter. 

Subsequently, we employ trilinear interpolation by querying the latent volume $\tilde{\boldsymbol{V}}$ with the sampled point cloud $\mathcal{X}_t^{*}$ to obtain the latent features $\mathcal{\hat{X}}_t$. The coordinates of both $\mathcal{X}_t^{*}$ and $\mathcal{X}_t$ are preserved for upsampling and positional embedding. 


\subsection{Dual Latent Mamba Blocks}
\label{sec:4.4}
Although time-variant frequency emphasis helps refine point selection, directly applying a state space model to raw points in each timestep is computationally expensive, given the high dimensionality and unordered nature of point clouds~\cite{pc-unorder-structure,pc-unorder-structure2,pc-unorder-structure3}. To address this, we propose Dual Latent Mamba Blocks (DM-Block), which operates in a latent space and serializes the downsampled point set into a 1D sequence conducive to Mamba modeling. It is designed to preserve local neighborhood relationships through diverse space-filling curves and to capitalize on Mamba ability to handle long-range dependencies efficiently.




\vspace{4pt}
\noindent\textbf{Space-Filling Curve Serialization}: 
To improve the sequential modeling ability of DM-Block, we reorder the latent points using Hilbert and Z space-filling curves and their transposed versions (Trans-Hilbert and Trans-Z), maintaining spatial proximity in a sequence. 
This reordering preserves spatial proximity in the sequence, allowing DM-Block to better capture local correlations as neighboring points remain close in the serialized representation.
Specifically, space-filling curves are paths that traverse every point within a higher-dimensional discrete space while maintaining spatial proximity to a certain degree and can be mathematically defined as a bijective function $\phi: \mathbb{Z} \rightarrow \mathbb{Z}^3$ for the point cloud. Given a space-filling curve $\mathcal{C}$, the latent point cloud $\mathcal{\hat{X}}_t$ is reordered according to its coordinates, resulting in the serialized latent point cloud as follows:
\begin{equation}
    \mathcal{\hat{X}}_t^{c} =  \mathcal{C} (\mathcal{X}_t^{*}) \mathcal{\hat{X}}_t, \quad \text{where}~\mathcal{\hat{X}}_t^{c} \in \mathbb{R}^{M \times D}.
\end{equation}

\begin{figure}[t]
\begin{center}
\includegraphics[width=1.0\linewidth,height=0.9\linewidth]{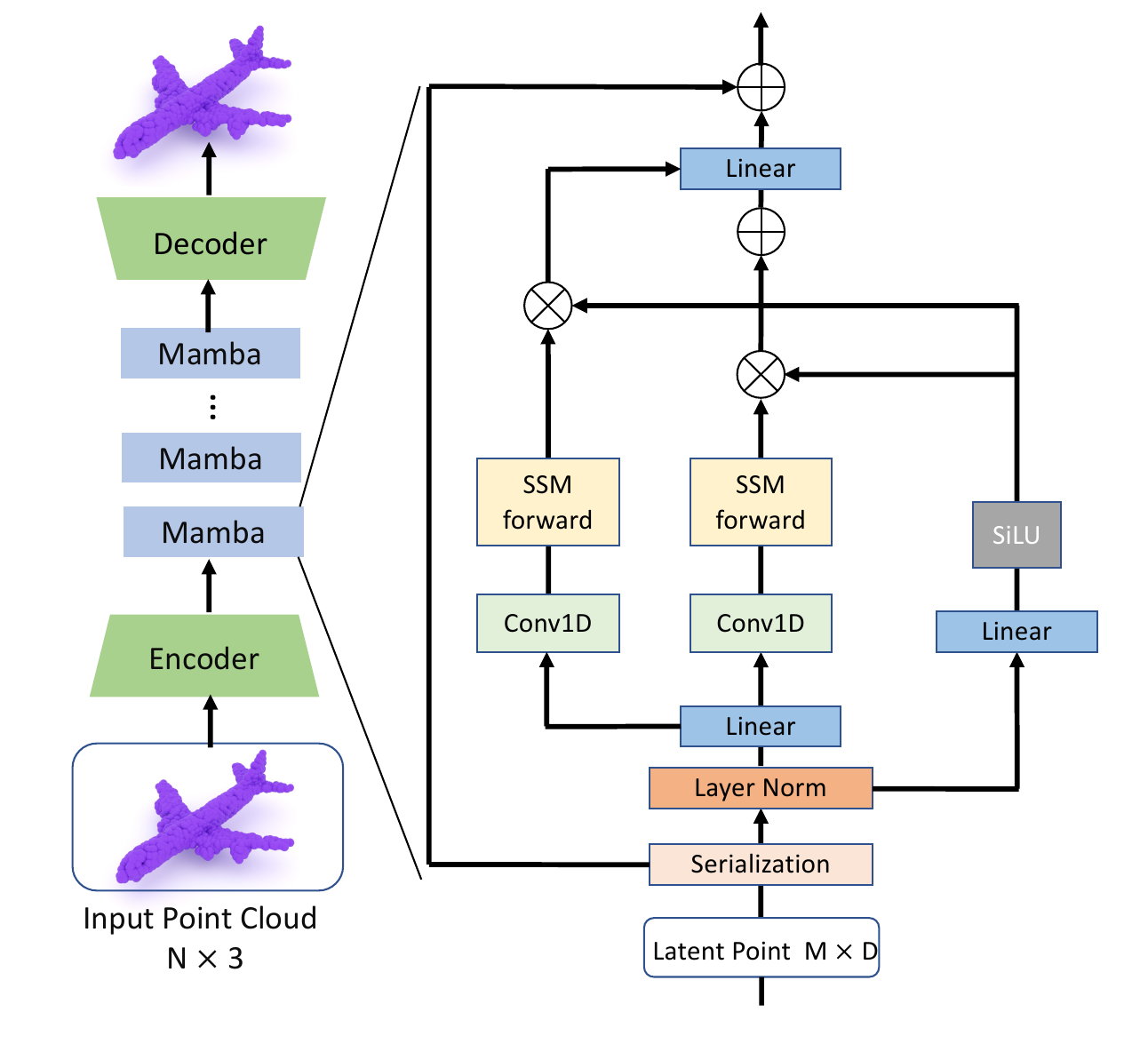}
 \vspace{-8mm}
\end{center}
   \caption{Illustration of our proposed Latent mamba block, which includes Layer Norm, Linear Layer, forward and backward state space model with its corresponding Conv1D block (N.B. we only perform serialization at the first block).}
\label{fig:mamba}
\end{figure}
\begin{table*}[htp]
\setlength{\tabcolsep}{10pt}
\centering
\resizebox{.95\textwidth}{!}{%
\begin{tabular}{|r|cc|cc|cc|cc|cc|cc|cc|}
\hline
\multirow{3}{*}{\textbf{Method}} & \multicolumn{4}{c|}{\textbf{Chair}} & \multicolumn{4}{c|}{\textbf{Airplane}} & \multicolumn{4}{c|}{\textbf{Car}} \\ \cline{2-13}
 & \multicolumn{2}{c|}{\textbf{1-NNA-Abs50 ($\downarrow $)}} & \multicolumn{2}{c|}{\textbf{COV ($\uparrow$)}} & \multicolumn{2}{c|}{\textbf{1-NNA-Abs50 ($\downarrow $)}} & \multicolumn{2}{c|}{\textbf{COV ($\uparrow$)}} & \multicolumn{2}{c|}{\textbf{1-NNA-Abs50 ($\downarrow $)}} & \multicolumn{2}{c|}{\textbf{COV ($\uparrow$)}} \\ \cline{2-13}
 & \textbf{CD} & \textbf{EMD} & \textbf{CD} & \textbf{EMD} & \textbf{CD} & \textbf{EMD} & \textbf{CD} & \textbf{EMD} & \textbf{CD} & \textbf{EMD} & \textbf{CD} & \textbf{EMD} \\ \hline
r-GAN~\cite{pc_GANs_rGAN}
& 33.69 & 49.70 & 24.27 & 15.13
& 48.40 & 46.79 & 30.12 & 14.32
& 44.46 & 49.01 & 19.03 & 6.539 \\

r-GAN (CD)~\cite{pc_GANs_rGAN}
& 18.58 & 33.94 & 19.71 & 15.43
& 37.30 & 43.95 & 38.52 & 21.23
& 16.49 & 38.78 & 38.92 & 23.58 \\

r-GAN (EMD)~\cite{pc_GANs_rGAN}
& 21.90 & 14.65 & 38.07 & 44.04
& 39.49 & 26.91 & 38.27 & 38.52
& 21.16 & 16.19 & 37.78 & 45.17 \\

PointFlow~\cite{pc_flow_pointflow}
& 12.84 & 10.40 & 46.84 & 47.35
& 25.68 & 20.74 & 47.04 & 40.52
& 8.10 & 6.52 & 35.40 & 44.60 \\

SoftFlow~\cite{pc_flow_softflow}
& 9.21 & 10.55 & 41.39 & 47.43
& 26.05 & 15.80 & 46.24 & 40.25
& 18.58 & 15.98 & 36.34 & 45.25 \\

SetVAE~\cite{pc_vae_setvae}
& 8.84 & 10.57 & 46.83 & 44.36
& 24.80 & 15.65 & 48.10 & 40.35
& 13.04 & 15.53 & 40.99 & 46.59 \\

DPF-Net~\cite{pc_flow_DPF-Net}
& 12.00 & 8.03 & 42.08 & 41.75
& 16.53 & 5.44 & 45.82 & 46.55
& 12.53 & 4.48 & 45.85 & 48.56 \\ \hline

DPM~\cite{pc_ddpm_dpm}
& 10.05 & 24.77 & 44.86 & 35.50
& 26.42 & 36.91 & 48.64 & 33.83
& 18.89 & 29.97 & 44.03 & 34.94 \\

PVD~\cite{pc_ddpm_pvd}
& 7.89 & 23.68 & 40.66 & 42.71
& 16.44 & 26.26 & 47.34 & 42.15
& 4.55 & 3.83 & 41.19 & 50.56 \\

LION~\cite{pc_ddpm_lion}
& 3.70 & 2.34 & 48.94 & 52.11
& 17.41 & 11.23 & 47.16 & 49.63
& 3.41 & 1.14 & 50.00 & \underline{56.53} \\

DiT-3D~\cite{pc_ddpm_DIT-3D}
& \textbf{0.89} & \textbf{0.73} & \textbf{52.45} & \underline{54.32}
& \textbf{12.35} & \underline{8.67} & \textbf{53.16} & \textbf{54.39}
& \underline{1.76} & 0.65 & 50.00 & 56.38 \\
FrePolad~\cite{pc_ddpm_freq}
& 3.53 & 3.23 & \underline{50.28} & 50.93
& \underline{15.25} & 12.10 & 45.16 & 47.80
& \textbf{1.89} & \underline{0.26} & \underline{50.14} & 55.23 \\

TIGER~\cite{pc_ddpm_tiger}
& 4.61 & 2.71 & - & -
& 21.85 & \textbf{5.82} & - & -
& 4.31 & 2.24 & - & - \\ \hline

{\ourmodel} (ours)
& \underline{3.25} & \underline{1.68} & 49.84 & \textbf{54.98}
& 18.31 & 8.88 & \underline{51.38} & \underline{52.25}
& 4.21 & \textbf{0.14} & \textbf{50.56} & \textbf{57.90}  \\ \hline

\end{tabular}%
}
\caption{\textls[0]{Comparison results (\%) on ShapeNet-v2 with shape metrics: Absolute 50-Shifted 1-Nearest Neighbor Accuracy (1-NNA-Abs50) and Convergence (COV), Chamfer Distance (CD) and Earth Mover’s Distance (EMD) are the distance metrics, where CD is multiplied by $10^3$ and EMD is multiplied by $10^2$; – denotes missing result due to unavailability from original authors; \textbf{Best}/\underline{2nd best} highlighted.}}
\label{tab:comp}
\end{table*}


\vspace{4pt}
\noindent
\textbf{Bidirectional Latent Mamba}:
For better efficiency and expressiveness, we employ a bidirectional variant of Mamba to capture forward and backward dependencies along the serialized sequence of the serialized latent point cloud $\mathcal{\hat{X}}_t^{c}$. Specifically, for each Mamba block, as shown in \cref{fig:mamba}, layer normalization~\cite{layer_norm}, causal one-dimensional convolution, SiLU activation~\cite{silu}, and residual connections are employed. The serialized latent point cloud sequence $\mathcal{\hat{X}}_t^{c}$ is processed through multiple Mamba blocks. Given an input $\mathcal{Z}_{l-1}$, the transformation in each block can be expressed as:
\begin{align}
   {\mathcal{Z}^{l}_{l-1}} &= \text{LN}(\mathcal{Z}_{l-1}), \notag \\
   {\mathcal{Z}^{'}} & = s(\text{Linear}(\mathcal{Z}^{l}_{l-1})),  \notag\\
   {\mathcal{Z}^{f}_l} & = \text{SSM}_\text{forward}(\text{Conv1D}(\text{Linear}(\mathcal{Z}^{l}_{l-1}))), \notag\\
   {\mathcal{Z}^{b}_l} & = \text{SSM}_\text{backward}(\text{Conv1D}(\text{Linear}(\mathcal{Z}^{l}_{l-1}))), \notag\\
   {\mathcal{Z}_{l}} & = \text{Linear}({\mathcal{Z}^{'}} \odot ({\mathcal{Z}^{f}_l}+{\mathcal{Z}^{b}_l}) ) + \mathcal{Z}_{l-1},
\end{align}
where $s$ represents the SiLU activation function, and ${\mathcal{Z}_{l}}$ is the output of the $l$-th block. The Mamba output $\mathcal{\hat{X}}_\text{out} \in \mathbb{R}^{M \times D}$ is obtained after passing stack of Mamba blocks.

\vspace{4pt}
\noindent
\textbf{Two Streams Affine Fusion}: 
To further enhance representation power, we run two parallel streams with different space-filling orders (e.g., Z vs. Z-Trans), each capturing distinct structural cues. We then propose to fuse them with a simple learnable affine transform, aligning features from both streams. This yields an aggregated representation that retains global shape coherence and local detail sensitivity.
Specifically, for the features output from different streams $\mathcal{\hat{X}}_\text{out}^{c1}$ and $\mathcal{\hat{X}}_\text{out}^{c2}$, we perform affine transformation as follows:
\begin{equation}
   \mathcal{\hat{X}}_\text{out}^{m} =  \text{Proj}((\mathcal{\hat{X}}_\text{out}^{c1} \odot \gamma^{c1} + \delta^{c1})  \oplus (\mathcal{\hat{X}}_\text{out}^{c2} \odot \gamma^{c2} + \delta^{c2})),
\end{equation}
\textls[-16]{where $\gamma^{c1},\gamma^{c2} \in \mathbb{R}^{D}$ and $\delta^{c1},\delta^{c2} \in \mathbb{R}^{D}$ are scale and shift factors, respectively. The operator $\odot$ denotes element-wise multiplication, and $\oplus$ denotes concatenation. $\text{Proj}(\cdot)$ represents a projection network that projects the concatenated features from  $\mathbb{R}^{M \times 2D}$to $\mathbb{R}^{M \times D}$. Subsequently, the final output feature $\mathcal{\hat{X}}_\text{out}^{m} \in \mathbb{R}^{M \times D}$ aggregates both global and local information.}

\begin{figure}[t]
\begin{center}
\includegraphics[width=1\linewidth]{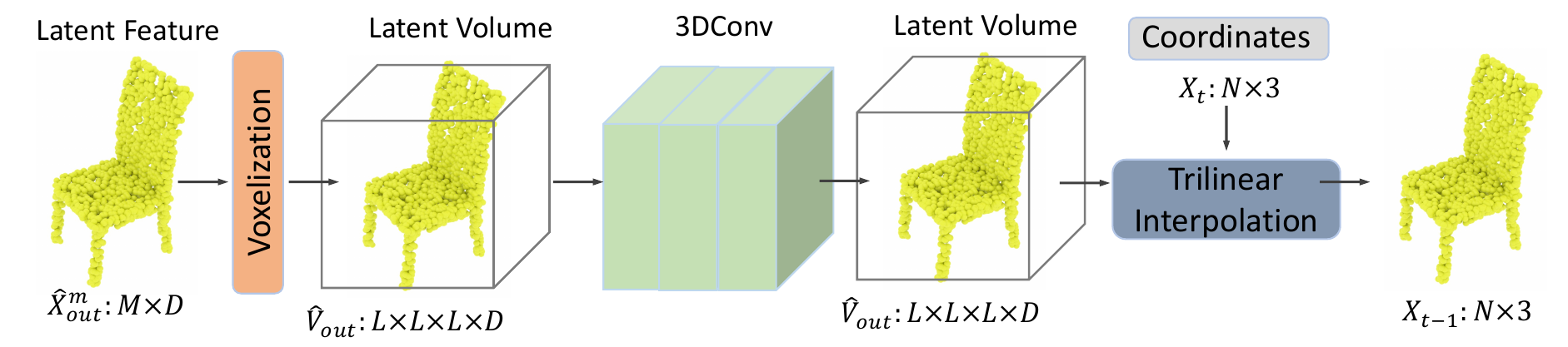} 
\end{center}
   \vspace{-5mm}
   \caption{The overview of decoder, the final prediction $X_{t-1}$ can be obtained by querying the latent volume $V_{out} $ with the coordinates.}
   \vspace{-2mm}
\label{fig:decoder}
\end{figure}

\vspace{4pt}
\noindent
\textbf{Point Cloud Decoder}: 
Finally, a point cloud decoder is applied to upsamples the latent point cloud to predict the noise $\mathbf{\epsilon}_\theta$, thus completing our diffusion pipeline. As shown in \cref{fig:decoder}, we employ trilinear interpolation to convert the latent point cloud $\mathcal{\hat{X}}_{out}^{m} \in \mathbb{R}^{M \times D}$ with the accompanying coordinates, to 3D space$\mathcal{X}_t \in \mathbb{R}^{N \times 3}$. Similarly to \cref{sec:4.3}, we voxelize the $\mathcal{\hat{X}}_{out}^{m}$ into volume $\tilde{\boldsymbol{V}_{out}} \in \mathbb{R}^{L \times L\times L \times D}$ and following an additional 3D convolutional network while preserving the original shape, then query use $\mathcal{X}_t$,  thereby obtaining the final prediction of the noise $\mathbf{\epsilon}_\theta$. By adopting the TF-Encoder and DM-Block, we not only overcome the computational bottleneck of raw 3D data processing but also retain high-frequency details at the correct diffusion phase. This holistic design integrates time-variant frequency emphasis with state space modeling in a straightforward yet novel manner.

\section{Experiments}
\label{sec:experiment}

We evaluate our proposed {\ourmodel} architecture against state-of-the-art 3D point cloud generation approaches on the established ShapeNet-v2~\cite{chang2015shapenet} benchmark dataset.

\vspace{-0.1cm}
\subsection{Experimental Setup}
\label{sec:exp-1}
\noindent
\textbf{ShapeNet-v2 Benchmark Dataset}: 
For a fair comparison on ShapeNet-v2, we follow the common practice that focuses on training and evaluating only select key categories, namely \textit{chair}, \textit{car}, and \textit{airplane}. From each shape, we sample 2048 points out of the 5000 available points in the training set and the test set, with normalization applied across the entire dataset. We adhere to the pre-processing steps and data split strategy as outlined in PointFlow~\cite{pc_flow_pointflow}.

\vspace{4pt}
\noindent
\textbf{Evaluation Metrics}:  
Following the popular practice of prior work~\cite{pc_ddpm_pvd,pc_ddpm_dpm}, we use 1-NNA (and the derived 1-NNA-Abs50) and COV to evaluate generation quality and diversity, alongside CD and EMD, which measure point-wise and distributional differences:
\begin{itemize}
    \item \textbf{1-NNA} (1-Nearest Neighbor) Accuracy: Measures the leave-one-out accuracy of a 1-NN classifier, reflecting both quality and diversity of generated samples.
    \item \textls[-18]{\textbf{1-NNA-Abs50} (Absolute 50-Shifted 1-NNA): Transforms the aforementioned 1-NNA $x$  into $|x-50|$, making it more sensitive to deviations from the ideal 50\%; a lower score indicates an ideal generated distribution closer to real data.}
    \item \textbf{COV} (Coverage): Evaluates how many reference point clouds are matched to at least one generated shape, where a higher value indicates greater diversity in generation.
    \item \textbf{CD} (Chamfer Distance): Measures point-wise similarity between generated and reference point clouds by computing the average nearest neighbor distance.
    \item \textbf{EMD} (Earth Mover’s Distance): Captures the minimal cost of transforming one distribution into another, providing a global similarity measure between point clouds.
\end{itemize}


\begin{table}
\setlength{\tabcolsep}{2pt}
 \begin{center}
    \resizebox{.48\textwidth}{!}
    {\begin{tabular}{@{}lccc|cccc@{}}
    
    \hline 
       & Serialization    & Freq     & Latent   & CD $\downarrow$  & EMD $\downarrow$ & CD  $\uparrow$  &  EMD $\uparrow$ \\
       & Strategy  & Decom & Block &(1-NNA-Abs50)   &(1-NNA-Abs50)  & (COV) & (COV)\\
    \hline
       (a) & None &    &   Conv   
       & 9.24   & 5.95  &   45.26 &  50.43   \\
       (b) & None      &           &   Transformer  
       & 6.21   & 1.43   &   49.65 &  54.15      \\
       (c) & Hilbert       &               &   Mamba     
       & 5.98   & 1.10   &   49.10 &  54.21       \\
       (d) & Hilbert        &  \checkmark   & Mamba       
       & 4.76   & 0.85   &   49.99 &  56.10       \\
       (e) & Hilbert + Hilber-Trans     &    &  Mamba      
       & 4.64   & 0.64   &   50.11 &  55.73       \\
       (f) & Hilbert + Hilber-Trans &  \checkmark & Mamba  
       & \underline{4.53}   & \underline{0.35}   & \underline{50.25}   & \underline{56.65}        \\
       (g) & Z + Z-Trans     &  \checkmark   &  Mamba      
       & \textbf{4.21}& \textbf{0.14}& \textbf{50.56} &  \textbf{57.90}       \\
    
    \hline
    \end{tabular}}
    \end{center}
    \vspace{-3mm}
    \caption{\textls[0]{Component-wise ablation of {\ourmodel} on ShapeNet-v2 (car category): {latent block,}serialization strategy, frequency-based component, and latent block. }
    }\label{tab:ablation}
\end{table}



\begin{figure}[t]
\begin{center}
\vspace{-3mm}
\includegraphics[width=0.95\linewidth,height=1.6cm]{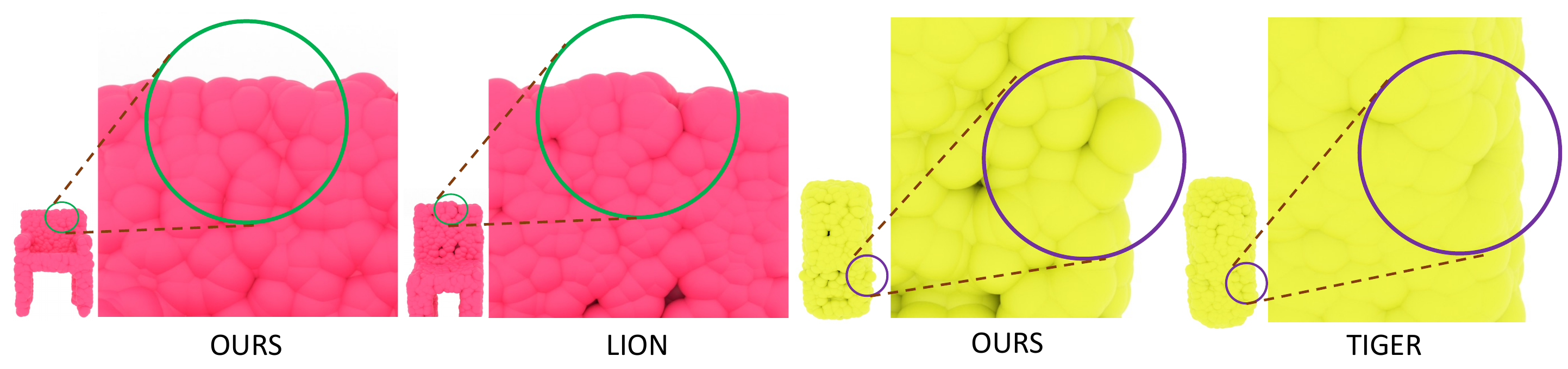}
\end{center}
   \vspace{-6mm}
   \caption{Back of chair (Pink) - smooth(ours)/deformed(other) , Car side-view mirror (Yellow) - retained(ours)/missing(other)}
   \vspace{-6mm}
\label{fig:details}
\end{figure}

\vspace{4pt}
\noindent \textbf{Implementation}: 
For the frequency-based encoder, we set $k=32$ for the k-NN graph, and the percentage $\zeta$ of high-pass points is set to 0.875. The diffusion model timestep is set to 1000, and the threshold $\tau$ for the sampling strategy is 50. In the Mamba layer, we apply 8 Mamba blocks for each stream with a latent size of 512. To enhance computational efficiency, the Mamba layer is applied to only 256 latent points. For training, we used an NVIDIA A100 GPU (80 GB) to train each category for 10,000 epochs with a batch size of 32. The learning rate was set to 0.0002, and we employed an Adam optimizer with a weight decay of 0.98.

\vspace{-0.1cm}
\subsection{Comparison with State-of-the-Art} 

\noindent
\textbf{Performance}: \textls[-2]In \cref{tab:comp}, we compare {\ourmodel} with multiple point cloud generation approaches. Notably, TIGER~(CVPR24)~\cite{pc_ddpm_tiger}, FrePolad~(ECCV24)~\cite{pc_ddpm_freq}, and DiT-3D~(NeurIPS23)~\cite{pc_ddpm_DIT-3D} are very recent methods. Among these, TIGER and FrePolad are relatively lightweight, yet we surpass both on the \textit{chair} and \textit{car} categories: for example, {\ourmodel} achieves a 0.25\% improvement in 1-NNA CD and 0.98\% improvement in 1-NNA-abs EMD on \textit{chairs} compared to the better of the two, and a 0.12\% gain on \textit{cars}. DiT-3D, while offering strong performance, incurs extremely high computational overhead, requiring 1700 GPU hours and 711 million parameters. Even so, {\ourmodel} outperforms it on three out of four metrics for the \textit{car} category, including a 0.51\% gain in 1-NN EMD and 1.52\% in COV EMD. These results highlight the efficiency and effectiveness of our approach.

\noindent
\textbf{Efficiency}: {As shown in \cref{tab:time}, our full model achieves the best results, requiring only slightly more training time than TIGER. Furthermore, compared to other top-performing methods, our approach significantly reduces training hours and parameter size while still achieving the highest overall performance. For further efficiency improvement, our single-stream variant achieves the lowest computational cost and fastest inference time, albeit with a slight performance trade-off compared to DiT-3D and our full model.}

\begin{figure}[t]
\begin{center}
\includegraphics[width=1\linewidth]{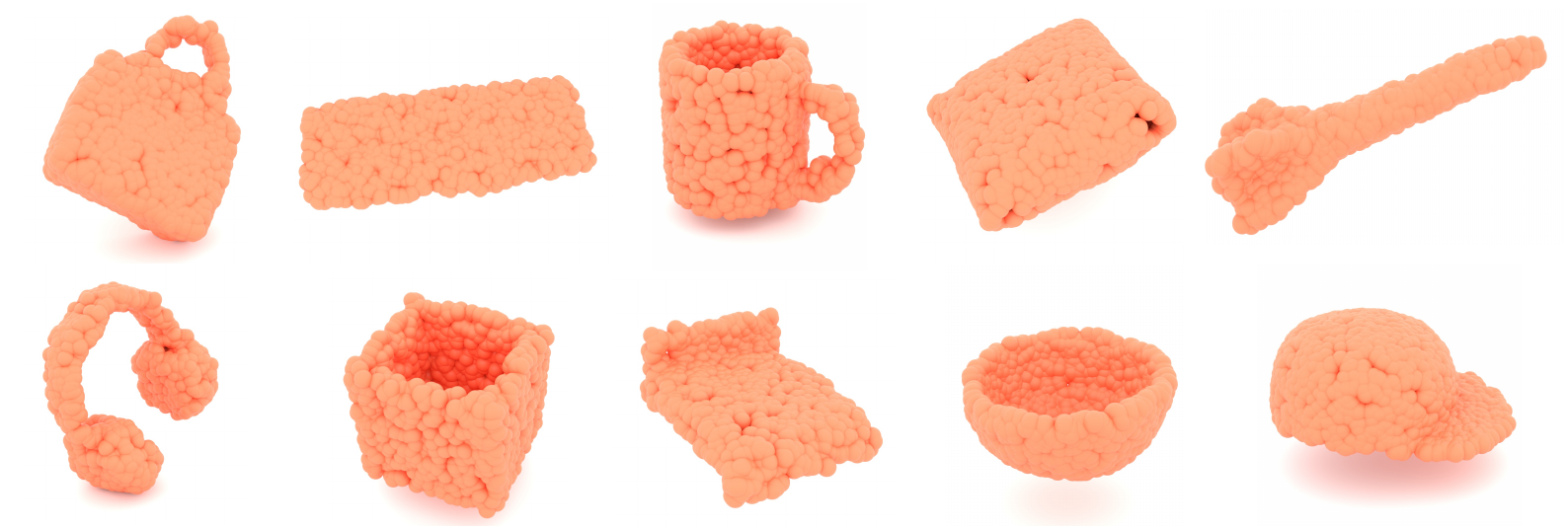} 
\end{center}
   \vspace{-5mm}
   \caption{Qualitative results of our model jointly trained on ten categories, presented in the following order: \textit{bag, keyboard, mug, pillow, rocket, earphone, basket, bed, bowl, and cap}.}
   \vspace{-4mm}
\label{fig:multi-gene}
\end{figure}

\noindent
\textls[-10]{{\textbf{Multi-Class Generation:} We train the TFDM model jointly without category conditioning on 10 object classes from ShapeNet-v2 (\textit{cap, keyboard, earphone, pillow, bag, rocket, basket, bed, mug, bowl}). Training on such a diverse set of shapes poses significant challenges due to the complexity and multimodal nature of the data. We present both qualitative (\cref{fig:multi-gene}) and quantitative (\cref{tab:multi-gene}) results. For comparison, we also train several baseline models under the same conditions, and the results demonstrate that our approach achieves the best overall performance across these methods.}}

\begin{table}[htbp]
\centering
\scriptsize  
\setlength{\tabcolsep}{3pt}  

\begin{tabularx}{\linewidth}{@{}XXXXX@{}}  
    \hline
    Method & Para (M) & Training Time (h) & Inference Time (s) & EMD (1-NNA-Abs50) $\downarrow$ \\
    \hline
    TIGER  & \underline{70.11} & \underline{164}  & \underline{9.73}  & 2.24 \\
    DIT-3D & 711.88         & 1688          & 100.13         & \underline{0.65} \\
    LION   & 144.25         & 550           & 27.12          & 1.14 \\
    {Ours (SS)}  & \textbf{48.84} & \textbf{138} & \textbf{8.12} & {0.85} \\
    Ours (Full)  & {70.25} & {192} & {11.41} & \textbf{0.14} \\
    \hline
 \end{tabularx}
\vspace{-3mm}
\caption{Training time, {inference time,} model size and the corresponding evaluation results. For a fair comparison, we report these metrics on Nvidia V100 GPU with a batch size of 32. Training time {and inference time}, measured in GPU hours {and second respectively}, is averaged over three categories: chair, airplane and car. Where `SS' indicates single-stream model.} 
\label{tab:time}
\vspace{-3mm}
\end{table}

\subsection{Ablation Studies}
\textls[-0]{In this section, we analyze the impact of various components and strategies within our proposed {\ourmodel} framework.}

\vspace{4pt}
\noindent \textbf{Different combinations of serialization methods}: We further evaluate multiple combinations of serialization methods within the two-stream architecture to determine the most effective strategy for information flow between streams. Specifically, for the car category, the combination of $z$ and z-transform serialization yields better performance than the combination of Hilbert and Hilbert transform. We compare row (f) with (e) in~\cref{tab:ablation}, the combination of z and z-trans order performs better than another, which gets improvements of 0.32\% in 1-NNA-Abs50 CD, 2.08\% in 1-NNA-Abs50 EMD, 0.31\% in COV CD, and 1.35\% in COV EMD.

\begin{figure}[t]
\begin{center}
\includegraphics[width=1.0\linewidth]{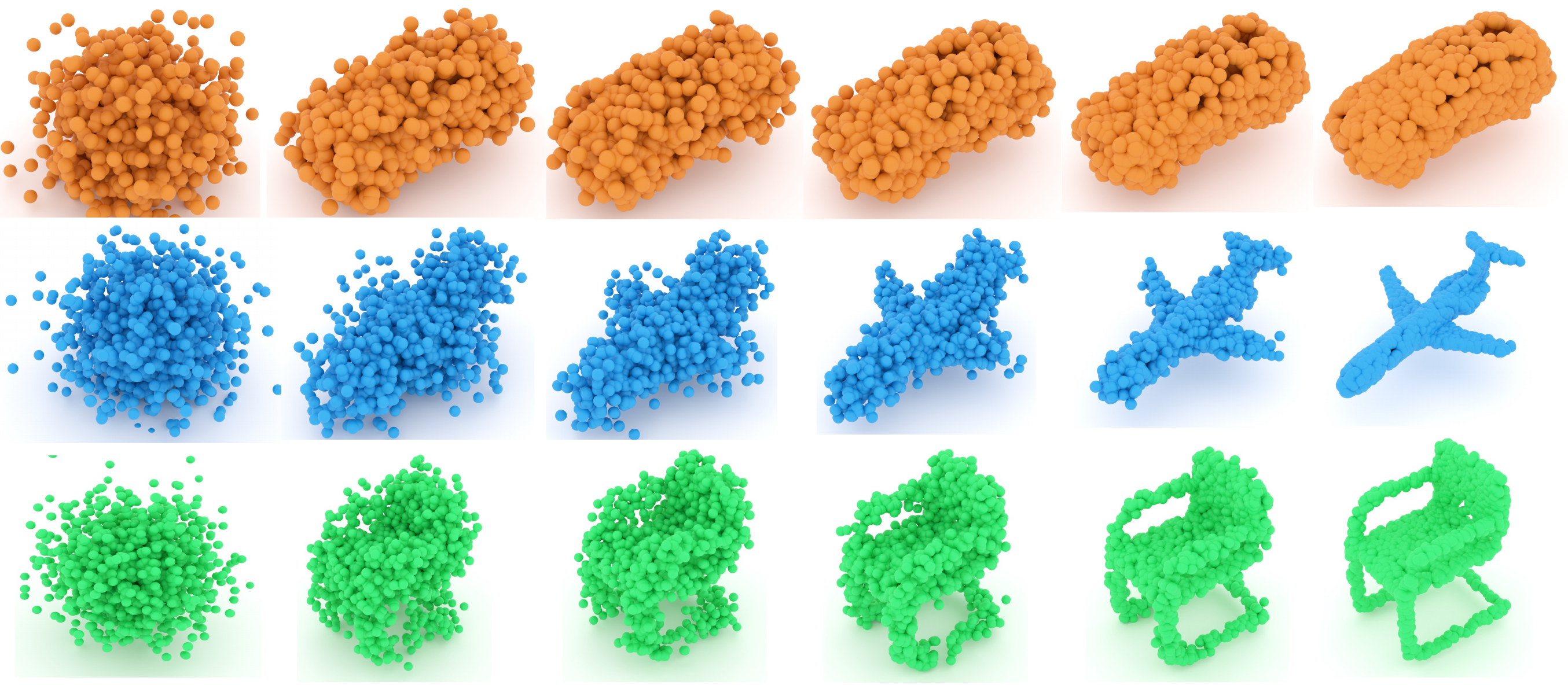}
\end{center}
   \vspace{-6mm}
   \caption{Illustrative examples of the reverse diffusion process demonstrating  detailed information recovery at the final timesteps (left to right, timesteps progressing from $T$ to 0). }
   \vspace{-2mm}
\label{fig:trace}
\end{figure}

\vspace{4pt}
\noindent \textbf{Effectiveness of frequency based model}: 
We assess the impact of the frequency-based time-variant strategy by comparing models that incorporate this mechanism against those that do not. Our results reveal that incorporating frequency decomposition leads to further improvements across all metrics. Specifically, comparing rows (d) and (e) in~\cref{tab:ablation}, the frequency-based method achieves improvements of 0.09\% in 1-NNA-Abs50 CD, 0.71\% in 1-NNA-Abs50 EMD, 0.13\% in COV CD, and 1.08\% in COV EMD. Additionally, as shown in~\cref{fig:trace}, the diffusion model recovers finer details in the final timesteps, making it well-suited for frequency analysis.

\vspace{4pt}
\noindent \textbf{Effectiveness of Mamba block}: In~\cref{tab:ablation}, comparing rows (a) and (b), we substitute the Mamba latent block (row, c) with standard 3D convolutional blocks (row, a) and Transformer block (row,b). The results demonstrate that substituting simple convolutional blocks with Mamba blocks significantly enhances both the quality and diversity of the generated outputs. Specifically, the Mamba blocks outperform the convolutional model by 3.26\% and 4.85\% in 1-NNA-Abs50 CD and EMD, respectively. Additionally, the Mamba block surpasses the Transformer block with 0.33\% in 1-NNA-Abs50 EMD and 0.55\% in COV CD while only containing half parameters of the Transformer block under the same conditions.

\begin{table}
\setlength{\tabcolsep}{10pt}
 \vspace{-2mm}
 \begin{center}
    \resizebox{.48\textwidth}{!}
    {\begin{tabular}{@{}c|cccc@{}}
    \hline 
        Method  & CD $\downarrow$  & EMD $\downarrow$ & CD  $\uparrow$  &  EMD $\uparrow$ \\
        &(1-NNA-Abs50)   &(1-NNA-Abs50)  & (COV) & (COV)\\
    \hline
        DPM &  9.71   & 21.54   &   43.65 &  38.94       \\
        PVD &  7.52   & 17.43   &   44.12 &  44.32       \\
        Tiger & 0.88  &  0.98   &   56.25  &  57.64       \\
        Ours &  \textbf{0.85}   & \textbf{0.43}   & \textbf{56.41}   &  \textbf{ 60.68 }    \\ 
      
    \hline
    \end{tabular}}
    \end{center}
    \vspace{-6mm}
    \caption{Comparison results (\%) jointly trained on ten categories.}
    \label{tab:multi-gene}
    \vspace{-4.5mm}
\end{table}

\vspace{4pt}
\noindent \textbf{Effectiveness of the two-stream Mamba layer design}: In~\cref{tab:ablation}, we evaluate the two-stream (row, d) versus the single-stream (row, b) architecture. \cref{tab:ablation} demonstrates our two-stream architecture consistently achieves superior results compared to the single-stream, regardless of other components. Our two-stream design achieves improvements of 1.34\% in 1-NNA-Abs50 CD, 1.09\% in 1-NNA-Abs50 EMD, 1.01\% in COV CD, and 1.52\% in COV EMD.

\vspace{4pt}
\noindent {\textbf{Effectiveness of Hyperparameters}: We also evaluate the impacts of hyperparamters $\tau$ and $\zeta$, As shown in \cref{tab:table4} for chair category, the results indicates that $\tau=50$ and $\zeta=0.875$ yield the best performance. The complete results are provided in the Supplementary Material.}


\begin{table}
\setlength{\tabcolsep}{10pt}
 \vspace{-2mm}
 \begin{center}
    \resizebox{.48\textwidth}{!}
    {\begin{tabular}{@{}lcc|cccc@{}}
    \hline 
       & $\tau$    & $\zeta$    & CD $\downarrow$  & EMD $\downarrow$ & CD  $\uparrow$  &  EMD $\uparrow$ \\
       &  &  &(1-NNA-Abs50)   &(1-NNA-Abs50)  & (COV) & (COV)\\
    \hline
      
       (a) & 25       &      0.875       
       & 3.54   & 1.99   &   49.01 &  53.99       \\
       (b) & 50  &  0.875      
       & 3.25   &  1.68   &   49.84  &  54.98       \\
       (c) & 50 &  0.75   
       & 4.15   & 2.37   & 48.93   &    53.46     \\ 
      
    \hline
    \end{tabular}}
    \end{center}
    \vspace{-6mm}
    \caption{Ablations on hyperparameters $\tau$ and $\zeta$ v.s. 1-NNA/COV.}
    \label{tab:table4}
    \vspace{-4.5mm}
\end{table}

\vspace{-.1cm}
\section{Conclusion}

\textls[-10]{In this paper, we propose a novel architecture that jointly leverages state-space models and frequency analysis within a point cloud diffusion framework for generative tasks. Our proposed DM-Block integrates latent space representations in the Mamba block to effectively address the challenge of efficiently applying diffusion via Mamba. Furthermore, we recognize that the diffusion process should recover fine-grained details during the final time steps. To this end, we introduce TF-Encoder, which includes a time-variant frequency-based point extraction method that achieves this without incurring high computational costs. Experimental results demonstrate that our method achieves state-of-the-art performance in certain categories while maintaining computational efficiency (with up to 10$\times$ less parameters and 9$\times$ shorter inference time than competitive approaches), and yields promising results across all categories.}

\noindent
\textbf{Future Direction}: Our method effectively achieves high-quality and diverse point cloud generation through the integration of frequency analysis and Mamba in latent space. However, our current approach applies frequency analysis without a dynamic adaptation mechanism. A promising direction would be to integrate frequency analysis directly with the neural network for joint training. 

\newpage
{
    \small
    \bibliographystyle{ieeenat_fullname}
    \bibliography{main}
}


\end{document}